\pdfoutput=1

\documentclass[11pt]{article}
\usepackage{authblk}

\usepackage[]{EACL2023}

\usepackage{booktabs}
\usepackage{graphicx}
\usepackage{enumitem}
\usepackage{xcolor}
\usepackage{soul}
\usepackage{multicol}
\usepackage{amsmath}
\usepackage{bold-extra}
\usepackage{subcaption}
\usepackage{float}
\usepackage{nicefrac}
\usepackage{color, colortbl}
\usepackage{times}
\usepackage{latexsym}
\usepackage{amssymb} 
\usepackage{multirow}
\usepackage{tablefootnote}
\usepackage{cleveref}
\usepackage{textcomp}
\usepackage{amssymb}
\usepackage{pifont}
\newcommand{\cmark}{\ding{51}}%
\newcommand{\xmark}{\ding{55}}%

\crefformat{section}{\S#2#1#3} 
\crefformat{subsection}{\S#2#1#3}

\renewcommand{\footnotesize}{\fontsize{8.5pt}{10.5pt}\selectfont}

\usepackage[T1]{fontenc}
\usepackage[utf8]{inputenc}

\usepackage{microtype}

\title{Towards a Unified Model for Generating Answers and Explanations \\ in Visual Question Answering}

\author{Chenxi Whitehouse, Tillman Weyde, Pranava Madhyastha \\
       City, University of London \\
       \texttt{\{chenxi.whitehouse, t.e.weyde, pranava.madhyastha\}@city.ac.uk} \\}

\begin{document}

\maketitle
\begin{abstract}

The field of visual question answering (VQA) has recently seen a surge in research focused on providing explanations for predicted answers.
However, current systems mostly rely on separate models to predict answers and generate explanations, leading to less grounded and frequently inconsistent results.
To address this, we propose a multitask learning approach towards a \textbf{U}nified \textbf{M}odel for 
\textbf{A}nswer and \textbf{E}xplanation generation (UMAE). 
Our approach involves the addition of artificial prompt tokens to training data and fine-tuning a multimodal encoder-decoder model on a variety of VQA-related tasks. 
In our experiments, UMAE models surpass the prior state-of-the-art answer accuracy on A-OKVQA by 10$\sim$15\%, show competitive results on OK-VQA, achieve new state-of-the-art explanation scores on A-OKVQA and VCR, and demonstrate promising out-of-domain performance on VQA-X.\footnote{Code is available at: \url{https://github.com/chenxwh/UMAE}.}

\end{abstract}

\section{Introduction}

Contemporary models for visual question answering (VQA) and commonsense reasoning are typically trained discriminatively to select the best answers from Multiple-Choice questions or to classify single-word answers to a predetermined vocabulary \citep[e.g.][]{anderson2018bottom}.
Such settings often lead to limitations such as encouraging models to find superficial correlations \cite{ye2021case} or penalising model performance even when the answers are plausible (e.g. synonyms and multi-word expressions, and morphological variations are not considered correct).
Most current explanation generation models are trained independently of the QA model and the explanations are usually generated after the QA model has provided an answer. 
As a result, these explanation models lack access to the process that generated the answer and thus the grounding of the explanation is limited to the answer text.

\begin{figure*}[!t]
\centering
    \includegraphics[width=0.95
    \linewidth]{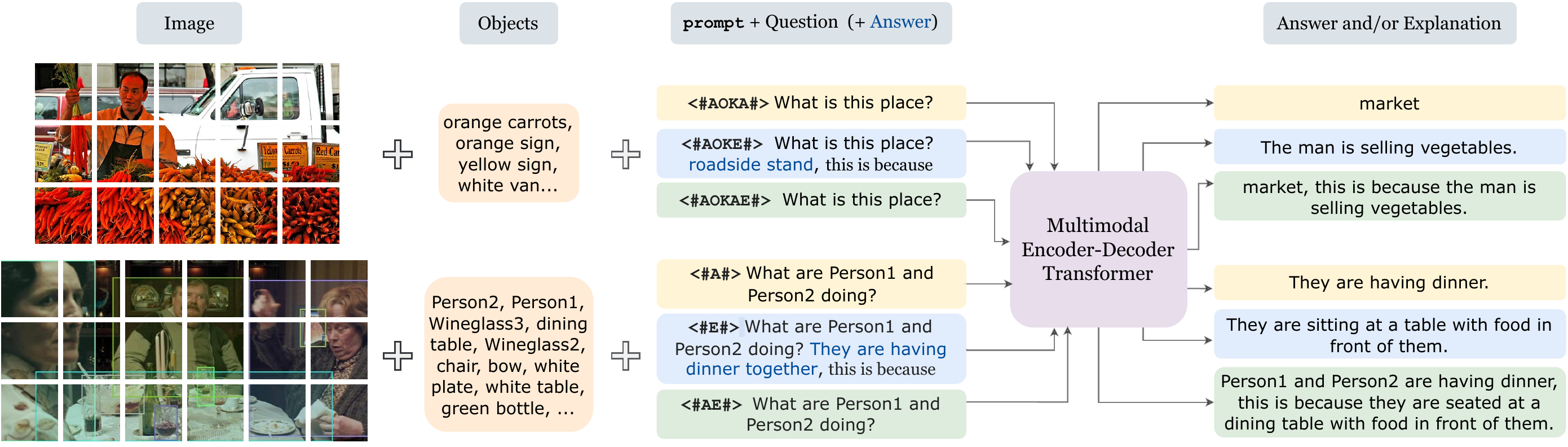}
    \caption{
    Illustration of UMAE:
    we train a multimodal encoder-decoder model on the mix of VQA tasks for jointly optimising answer and explanation, where we distinguish the training instances and target output with artificial prompt tokens (e.g. \texttt{<\#AOKA\#>}).
    The top and bottom examples are from A-OKVQA and VCR, respectively.
    }
\label{fig:illustration}
\end{figure*}

We posit that a unified model that simultaneously performs answer prediction and explanation generation is a more effective and consistent approach for VQA.
Generative models, such as GPT-3 \cite{brown2020language}, T5 \cite{JMLR:v21:20-074}, or OFA \cite{Wang2022UnifyingAT}, have been shown to be successful at rapidly adapting to downstream tasks and generating high-quality open-ended text, and hence are suitable candidates for this unified approach.

We propose a multitask learning approach for multimodal transformer-based encoder-decoder models, towards a United Model for Answer and Explanation generation (UMAE).
In addition to the current trend of separate answer prediction and explanation generation based on the answers, our approach adds the capability of jointly generating answers and explanations together. 
Inspired by the success of artificial prompt tokens in Neural Machine Translation (NMT) \cite{johnson-etal-2017-googles}, we extend and demonstrate the efficacy of the artificial prompt-based method for VQA in a multitask setup. 
We augment training instances with artificial prompt tokens, enabling the model to distinguish different tasks while learning shared semantic features. 
Experiments on a combination of three knowledge-intensive VQA datasets, OK-VQA \cite{marino2019ok}, A-OKVQA \cite{Schwenk2022AOKVQAAB}, and VCR \cite{zellers2019vcr}, show that the UMAE models achieve a new state-of-the-art (SOTA) answer accuracy on A-OKVQA, new SOTA explanation score on VCR, and competitive out-of-domain performance on VQA-X \cite{park2018multimodal}.
UMAE supports the generation of the answer to a question, the explanation for a given question and answer, and both together jointly, making the model efficient and flexible. 
An illustration of the training setup is shown in \autoref{fig:illustration}.

Our main contributions are:
a) the UMAE framework where answers and explanations can be generated by a single unified model (\S\ref{sec:umae});
b) a simple and efficient training approach that uses multitask learning with artificial prompts and demonstrates its ability to generalise across domains (\S\ref{sec:setup});
c) a method to map generated answers to Multiple-Choice options via evaluating the perplexity of the generation (\S\ref{sec:ppl});
d) new SOTA results by UMAE, particularly for explanation generation and promising out-of-domain performance (\S \ref{sec:results}).

\section{Related Work} \label{sec:related_work}
\noindent
\textbf{Multimodal Transformer-based Models} achieve SOTA performance on various vision-language tasks \cite{chen2020uniter, li2020oscar, cho2021unifying, wang2022simvlm, zhang2021vinvl}. 
They showcase the possibility of capturing richer multimodal semantic coherence than discriminatively trained models and are further capable of generating self-explanations. 
Pretrained on multitask settings with natural language instructions, e.g. \textit{``what does the region describe?''}, models like OFA \cite{Wang2022UnifyingAT} are claimed to have the capability to transfer to unseen tasks and domains via similar instructions. However, contrary to these claims, we observe that pretrained OFA is incapable of generating valid explanations through simple natural language instructions (\cref{sec:results}). 

\vspace{0.2cm}
\noindent
\textbf{Artificial Prompt Tokens} have previously been explored for NMT by \citet{johnson-etal-2017-googles,mitzalis-etal-2021-bertgen}.
They propose a single model with the traditional NMT model architecture (usually for one language pair) and jointly train on different language pairs with added artificial prompts, e.g. \texttt{2es} to distinguish the target language.
This approach has been found to foster implicit cross-lingual bridging and exhibit zero-shot translation capability. 
In this paper, we exploit a similar approach with artificial prompts for answer and explanation generation in VQA with a united model. 
This enables the model to learn shared features among tasks and datasets in various domains.

\vspace{0.2cm}
\noindent
\textbf{Explanation Generation for VQA} has gained growing interest in research.
However, most recent approaches use separate models to predict answers and generate explanations \cite{dua2021beyond}.
\citet{wu-mooney-2019-faithful} generate explanations with an object detector and a GRU unit for text embedding, then train on a subset of VQA-X in which the explanations contain the objects most attended to by the model. 
\citet{kayser2021vil} develop an e-UG model combining UNITER \cite{chen2020uniter} for processing multimodal input and GPT-2 \cite{radford2019language} for generation. 
In contrast, in this paper, we propose using a single united model for more grounded answer and explanation generation.

\section{Methodology}
\label{sec:methodology}


\begin{table*}[t!]
\centering
\scalebox{0.75}{
\begin{tabular}{l|cccccccc}
\toprule
{\multirow{3}{*}{\sc Model}} & {\sc Ok-vqa} & \multicolumn{5}{c}{\sc A-okvqa}  & \multicolumn{2}{c}{\sc Vcr}\\
& \textit{direct answer} & \multicolumn{3}{c}{\textit{multiple choice}} & \multicolumn{2}{c}{\textit{direct answer}}  & \textit{multiple choice} & \sc bertscore\\

& \sc test &\textsc{val} (\textit{ppl}) &\textsc{val} (\textit{GloVe})  &\sc test  & \sc val   & \sc test    & \textsc{val} (\textit{ppl}) & \sc val \\
\cmidrule(lr){1-1} \cmidrule(lr){2-2} \cmidrule(lr){3-5} \cmidrule(lr){6-7} \cmidrule(lr){8-9}
\sc ofa* & 40.40 & 24.54 &  56.19 & 47.40 & 48.09 & 39.77 & 33.55 & 64.55 \\
\sc ofa\textsubscript{q->a}  & 49.93 & 74.32 & 65.30   & 61.71 &  63.00  & 53.91  & 54.89 & 83.85 \\

\sc umae\textsubscript{all}  & \textbf{51.77} & \textbf{74.59} &  \textbf{65.67}   & \textbf{63.26}  & \textbf{63.29}   & \textbf{56.14}  & \textbf{56.66} &\textbf{85.97} \\
\midrule
\sc Prior-best  & 54.41 &  -- &60.30 & 53.70  & 48.60  & 40.70 & (77.10)\rlap{\textsuperscript{\dag}} & --  \\
\bottomrule
\end{tabular}
}
\caption{Performance of models for answer generation.
Better results are in bold.
\textsc{ofa}* refers to the pretrained OFA.
Prior-best results
for the three datasets 
are from \citet{gui-etal-2022-kat}, \citet{Schwenk2022AOKVQAAB}, \citet{wang2022vqa}, respectively.
\dag{} is from a discriminative model and thus not comparable 
\citep[see][]{ye2021case}.
}
 \label{tab:ans_acc}
\end{table*}

\begin{table*}[!ht]
\centering
\scalebox{0.75}{
\addtolength{\tabcolsep}{0pt}
\begin{tabular}{ll|rrrrrrrrc}
\toprule
{\multirow{2}{*}{\sc Dataset}}
 & {\multirow{2}{*}{\sc Model}} & \multicolumn{3}{c} {e{\sc-V}i{\sc l Scores}} & \multicolumn{5}{c} {\sc {N-gram Scores}} & \sc Learnt Score
\\

 & & \sc \textit{S}\textsubscript{O}    & \sc \textit{S}\textsubscript{T}     & \sc \textit{S}\textsubscript{E}   & \multicolumn{1}{c}{\sc bleu\small{4}}  & \sc rouge-l   &\sc meteor  & {\sc cide}r & \sc spice & \sc bertscore    \\
\cmidrule(lr){1-2} 
\cmidrule(lr){3-5} \cmidrule(lr){6-10} \cmidrule(lr){11-11}

{\multirow{4}{*}{\sc {A-okvqa}}}
&\sc ofa* &   4.44
 & 56.19 & 
7.90			& 	0.30&	4.45& 	3.26	& 4.82	&4.62& 	68.64
\\
&\sc ofa\textsubscript{q->a}+ofa\textsubscript{qa->e} &   
35.82 & 74.32 &
48.29&	22.18&	48.51	&23.56	&86.76	&22.46	&	85.96
\\

& \sc umae\textsubscript{a-okvqa} & 37.10 & 73.97 & 50.15  & \textbf{27.61} & 52.23 & 24.06 & \textbf{104.39} & 22.88 & 87.86    \\

&\sc umae\textsubscript{all} & \textbf{37.91} & \textbf{74.59} &\textbf{50.82} & 27.35 & \textbf{52.56} & \textbf{24.83} & 101.09 & \textbf{23.33} & \textbf{88.21}    \\
\midrule
{\multirow{3}{*}{\sc {Vcr}}} 
& e{\sc-ug} &19.30 & \textbf{69.80} & 27.60  &4.30 & 22.50 & 11.80 & 32.70 & 12.60 & 79.00   \\

& \sc umae\textsubscript{vcr} & 22.57 & \textbf{56.68} & 39.82 & 12.25 & 28.87 & 16.67 & \textbf{48.14} & 27.36  & 81.77\\
& \sc umae\textsubscript{all} & \textbf{22.82} & 56.66 & \textbf{40.27} & \textbf{13.44} & \textbf{29.53} & \textbf{17.54} & 47.33 & \textbf{26.45}   & \textbf{81.91} \\

\midrule
{\multirow{2}{*}{\sc {Vqa-x}}}
&  e{\sc-ug} & 36.50 & 80.50 & 45.40 &  23.20 & 45.70 & 22.10 & 74.10 & 20.10 & 87.00\\


&\cellcolor[HTML]{DDEBF7}\sc umae\textsubscript{all} & \cellcolor[HTML]{DDEBF7}{31.58}  &\cellcolor[HTML]{DDEBF7}{77.65}   &\cellcolor[HTML]{DDEBF7}{40.67} & \cellcolor[HTML]{DDEBF7}{14.63} & \cellcolor[HTML]{DDEBF7}{35.12} & \cellcolor[HTML]{DDEBF7}{20.29} &\cellcolor[HTML]{DDEBF7}{50.35} &\cellcolor[HTML]{DDEBF7}{19.13}  & \cellcolor[HTML]{DDEBF7}{85.40}    \\

\bottomrule
\end{tabular}
}
\caption{Explanation Scores. 
\textsc{ofa}* is the pretrained OFA, showing the transferability of OFA for generating explanations with natural language instructions.
Results with e-UG are from \citet{kayser2021vil}.
We show the best results of A-OKVQA and VCR in bold.
The last row in blue shade shows \textit{out-of-domain} performance.
}
\label{tab:explanation}
\end{table*}

\subsection{Multitask Learning with Artificial Prompt}
\label{sec:umae}

We formulate three generation settings: \texttt{Q$\rightarrow$A}: answer prediction; \texttt{QA$\rightarrow$E}: explanation generation conditioned on the answer; and \texttt{Q$\rightarrow$AE}: 
\textit{joint} answer and explanation generation for a given question.
We hypothesise that by training the model to generate both the answer and its explanation \emph{simultaneously}, the result answer and explanation will be more grounded and consistent.

We use a pretrained multimodal encoder-decoder transformer as our base model (here we build on the openly released version of OFA as a strong baseline), and finetune the model on a mix of VQA datasets from different domains. 

Different from OFA, for each image in the VQA datasets, we first extract objects and attributes using a bottom-up top-down attention-based model, which is crucial for open-domain VQA tasks \cite{anderson2018bottom}. 
We then add artificial prompt tokens at the beginning of the textual input to signal the generation task (answer, explanation, or both) and the dataset\footnote{Artificial prompt tokens are added as special tokens to the tokenizer to avoid bias in the pretrained embeddings. However, we note that these tokens may be biased w.r.t their association with specific tasks after training, which is an intended effect.}.
For \texttt{Q$\rightarrow$AE}, we concatenate answers and explanations with a separator in between.
Finally, we mix all training instances, each consisting of an image (processed in patches), objects and attributes, and textual input with artificial prompts.

\subsection{Perplexity as Multiple Choice Metric}
\label{sec:ppl}
To map the generated output to Multiple-Choice options, in previous work the predictions are loosely matched with options or gold answers using embedding-based methods, such as GloVe embedding similarity \citep{Schwenk2022AOKVQAAB}. 
In contrast to these approaches, we propose to evaluate each option as a \textit{text generation} task, by feeding the model the information that was used to generate the answer as a prompt, and calculating the likelihood of each option being generated.
Formally, given an option $Y = (y_1, y_2, ..., y_t)$ with $t$ tokens, we calculate the probability of each token $y_i$ being generated by feeding the image, objects, and question, as well as the first $i-1$  tokens from $Y$ to the model $p_\theta$.
The perplexity is then calculated with: $\mathrm{PPL}(Y) = \mathrm{exp}\left\{ -\frac{1}{t}\sum_{i}^{t}\mathrm{log}\,p_{\theta}\left(y_{i}|y_{< i}  \right) \right\}$, which reflects the probability of option $Y$ being generated by the model.
Finally, the option with the lowest perplexity is chosen as the answer.

We also compare the performance of our approach, using perplexity as the metric, with GloVe embedding similarity for A-OKVQA (see \autoref{tab:ans_acc}).

\begin{figure*}[ht]
\centering
    \includegraphics[width=0.95\linewidth]{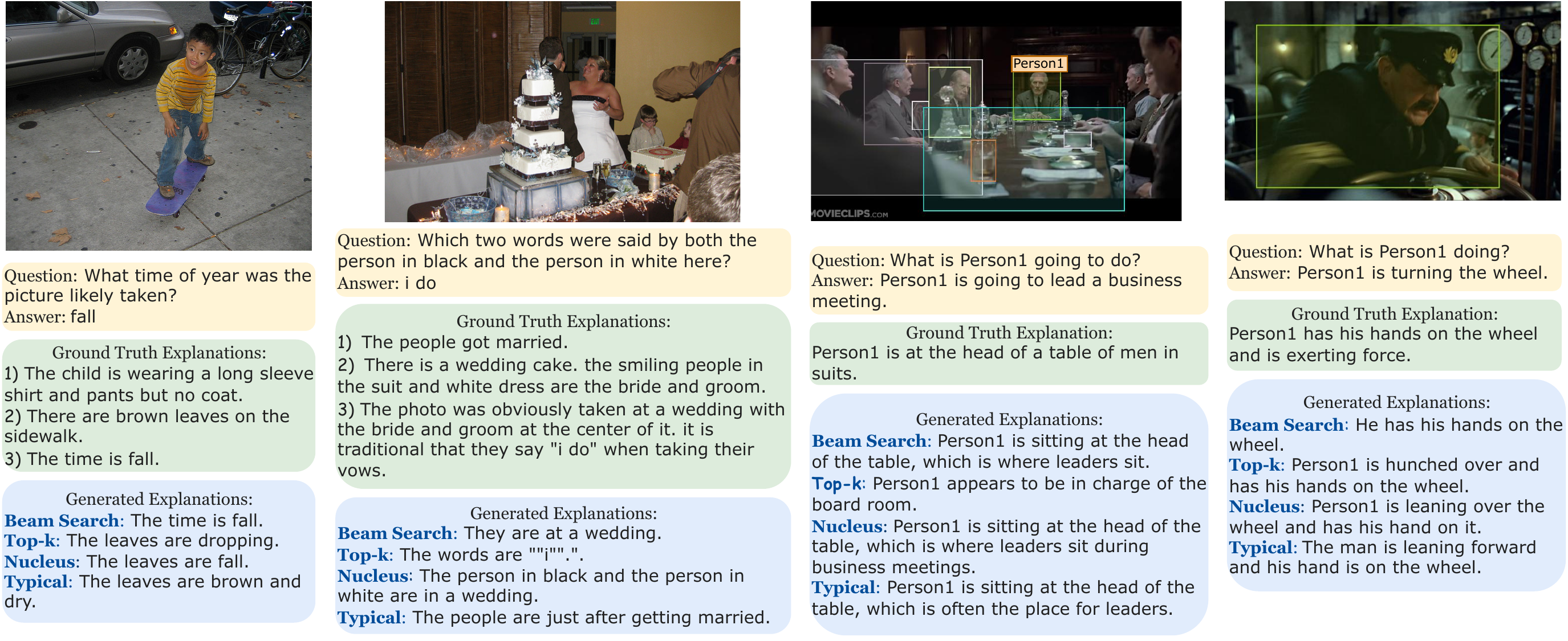}
    \caption{Examples of generated explanations from \textsc{umae\textsubscript{all}} model with different decoding strategies.
    The two examples on the left are from A-OKVQA and the two on the right are from VCR.}
\label{fig:explanations}
\end{figure*}

\section{Experimental Setup}
\label{sec:setup}

We primarily evaluated our proposed UMAE approach using pretrained OFA\footnote{\url{https://github.com/OFA-Sys/OFA}} as the base model on three knowledge-intensive VQA datasets: 
OK-VQA, A-OKVQA and VCR\footnote{See \autoref{sec:datasets} for datasets details.}.
We split the original train set into train and validation set (95\%-5\%) for all three datasets. Since the test set is not publicly available for A-OKVQA and VCR, we use the original validation set for experimental analyses. 
We prepare training instances\footnote{Specifically, we add \texttt{<\#OKA\#>} for OK-VQA (only answers are available), \texttt{<\#A\#>}, \texttt{<\#E\#>}, \texttt{<\#AE\#>} for VCR, and \texttt{<\#AOKA\#>}, \texttt{<\#AOKE\#>}, \texttt{<\#AOKAE\#>} for A-OKVQA.} as introduced in \cref{sec:umae}.
Additionally, for VCR, we draw coloured highlights around the referenced entity on the images, following \citet{NEURIPS2021_c6d4eb15} (\autoref{sec:datasets}).
To account for the imbalance in size among the datasets, we up-sample 
instances in OK-VQA and A-OKVQA, and shuffle all instances to train the \textsc{umae\textsubscript{all}} model.

For ablation studies, we finetune OFA for separate answer prediction (\textsc{ofa\textsubscript{q->a}}) and explanation generation conditioned on answers (\textsc{ofa\textsubscript{qa->e}}). 
To better understand the impact of mixing datasets from different domains, we also train \textsc{umae\textsubscript{a-okvqa}} and {\textsc{umae\textsubscript{vcr}}},  focusing on all three answer and explanation generation tasks but only using data from a single dataset: either with A-OKVQA or with VCR.
Details of training parameters are included in \autoref{sec:parameters}.

We use beam search for generating answers and additionally experiment with different decoding methods including top-k sampling, Nucleus sampling \cite{Holtzman2020The}, and Typical sampling \cite{meister2022typical}, for generating explanations.
We evaluate answer accuracy as well as explanation quality with automatic NLG metrics and e-ViL scores \cite{kayser2021vil}.
e-ViL scores consist of \texttt{S\textsubscript{T}} (task/answer accuracy),
\texttt{S\textsubscript{E}} (explanation score), and overall \texttt{S\textsubscript{O}} (product of \texttt{S\textsubscript{T}} and \texttt{S\textsubscript{E}}), where \texttt{S\textsubscript{E}} is the harmonic mean of NGRAMScore (the harmonic mean of n-gram scores ROUGE-L \cite{lin-och-2004-automatic}, METEOR \cite{banerjee-lavie-2005-meteor}, CIDEr \cite{ vedantam2015cider}, and  SPICE \cite{anderson2016spice}) and additionally the BERTScore \cite{bert-score}, a learned similarity metric over contextual representations of sentences.

\section{Results and Discussion}
\label{sec:results}

\subsection{Answer Accuracy}
\autoref{tab:ans_acc} presents our observations for answer accuracy on \texttt{Q->A} task over the three datasets. 
We also evaluate VCR answers using BERTScore as the answers for VCR are usually sentences.
We observe that \textsc{umae\textsubscript{all}} outperforms \textsc{ofa\textsubscript{q->a}} on all datasets, improves the prior SOTA on A-OKVQA by 10$\sim$15\%, and achieves competitive results on OK-VQA.
For models that are finetuned on A-OKVQA, we also see a salient improvement (+9\%) with the proposed mapping of options by  perplexity in Multiple-Choice, instead of GloVe embeddings similarity\footnote{Preliminary experiments with NLG metrics (BERTScore and BLEU) for selecting the options given generation were sub-optimal.}.
We conducted several ablation studies on the dependency of the modality for the answer accuracy in A-OKVQA, where we find the visual encoder is crucial for performance.
Details are included in \autoref{sec:modality}.

\subsection{Explanation Evaluation}
\autoref{tab:explanation} shows e-ViL sores (\cref{sec:setup}) for explanations using automatic NLG metrics\footnote{Nucleus sampling shows best results and is reported. 
Detailed scores
with different decoding methods are shown in \autoref{sec:explanation_scores}.}.
Following the same setup as in \citet{kayser2021vil}, an explanation is evaluated only if the answer predicted by the system is correct\footnote{A limitation of evaluating all explanations is that explanations of wrong answers may get high scores with n-gram metrics, even though they are justifying wrong answers and should be penalised.}.
We observe that pretrained OFA with natural language prompts, e.g. \textit{``what is the explanation for the answer?''} or \textit{``this is because''} performs poorly, as most 
generated explanations are words (\textit{``yes/no''}) or short-phrases\footnote{BERTScore in not representative of the validity of outputs from \textsc{OFA}*. We refer the reader to an exposition of the problems associated with NLG metrics in \citet{caglayan-etal-2020-curious}.}.
We compare UMAE models (on all and individual datasets) with prior best results from e-UG (see \cref{sec:related_work}), and standard separated trained baselines (\textsc{ofa\textsubscript{q->a}}+\textsc{ofa\textsubscript{qa->e}}).
\textsc{umae\textsubscript{all}} achieves better results across all datasets, showing the advantage of mixing tasks and datasets in different domains.
For out-of-domain evaluation on VQA-X, \textsc{umae\textsubscript{all}} also shows mostly competitive results.
Examples of explanation generation are shown in \autoref{fig:explanations} and \autoref{sec:explanations}.

Since e-ViL only evaluates an explanation if a model generates the correct answer, the subset of explanations evaluated varies by model.
To \textit{fairly} compare explanations on the same subset, we propose only using the subset of samples where all models provide correct answers for explanation prediction. 
\autoref{tab:exp} shows the results on A-OKVQA with such a subset of 770 candidates, where \textsc{umae\textsubscript{all}} shows an even higher explanation score. This highlights that \textsc{umae\textsubscript{all}} generates explanations that overlap significantly better with gold explanations.

\begin{table}[!t]
\centering
\scalebox{0.7}{
\addtolength{\tabcolsep}{-3.5pt}
\begin{tabular}{l|cccccccc}
\toprule
\sc Model & \sc \textit{S}\textsubscript{E}   & {\sc bleu\small{4}}  & \sc r-l   &\sc met.  & {\sc cide}r & \sc spice & {\sc berts}c. 
\\
\midrule
\textsc{ofa\textsubscript{q->a}+ofa\textsubscript{qa->e}}& 42.4 & 20.0	&44.2	&19.3&	66.7	&19.1	&	85.1\\

\sc {umae\textsubscript{a-okvqa}}& 45.8 &	 23.6&		47.9&		21.7	&	78.0	&	20.5	&	86.9\\
\sc {umae\textsubscript{all}} & \textbf{46.8}& \textbf{24.9}	& \textbf{49.5} &	\textbf{22.3}  &	\textbf{84.1}   &	\textbf{20.8}	&	\textbf{87.3} \\

\bottomrule
\end{tabular}
}
\caption{
Explanation scores on the same subset of A-OKVQA.
}
\label{tab:exp}
\end{table}

In summary, our experiments demonstrate that the UMAE model leads to improved answer and explanation generation and allows for the flexibility to generate different types of outputs, including answers, explanations, or both. We observe that UMAE exhibits promising results in jointly generating both the answer and explanation. 
We further provide a comparative evaluation in \autoref{sec:joint} as a first step towards comparison as there is currently no standard evaluation setup for the joint answer and explanation evaluation.

\subsection{Error Analysis}
\label{sec:analysis}

\begin{figure}[!t]
\centering

    \includegraphics[width=0.95\linewidth]{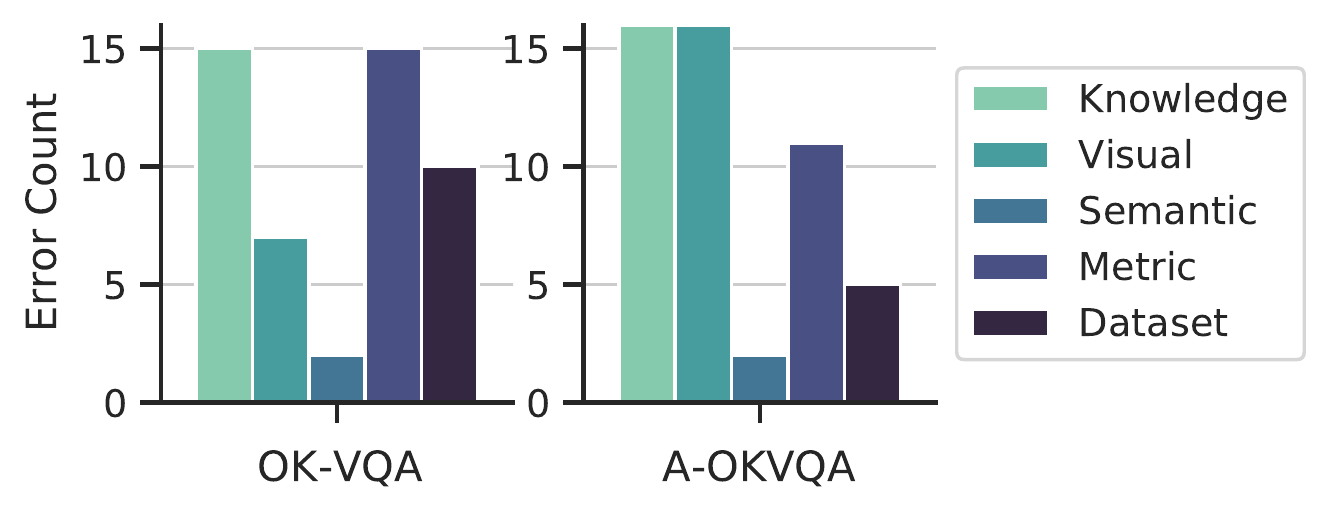}
    \caption{Error type distribution in 100 random samples from A-OKVQA and OK-VQA.
    }
\label{fig:errors}
\end{figure}

To better understand the generated answers and errors, we randomly sample 50 errors in OK-VQA and A-OKVQA.
Our analysis reveals the following main error types, where the first three are related to model performance:
(1) \textit{Knowledge}:  
the implicit knowledge learned by the model is insufficient for answering some of the knowledge-intensive questions, such as questions asking \emph{when} a certain sport was invented;
(2) \textit{Visual}: 
the model fails to identify the visual attributes correctly, such as questions about \emph{recognising object shape or material};
(3) \textit{Semantic disassociation}: 
the model misinterprets questions or fails to match the intended semantic meaning. For example, it may answer what \emph{an object is} instead of a more complex question such as \emph{what is commonly packed in it} (e.g. answering "suitcase" instead of "clothes");
(4) \textit{Metric}: 
the evaluation metric may penalise some of the plausible answers, especially when searching for exact match answers (mostly due to the difference of singular/plural or phrases with/without space in between);
and
(5) \textit{Dataset}:  
errors due to issues in the datasets themselves. We discuss prominent issues in dataset quality briefly in  \autoref{sec:datasets_issues} and further present the distribution of error types in \autoref{fig:errors}.

\section{Conclusions}
In this work, we propose UMAE, a unified model that generates answers and explanations in VQA using a multitask learning approach for multimodal encoder-decoder models, where artificial prompt tokens are added to distinguish different tasks while learning shared semantics.
Evaluation of our approach on various VQA tasks shows that UMAE outperforms prior best models and separately trained baselines in both answer and explanation scores, where we also demonstrate the benefit of using perplexity as the metric for mapping generated answers to Multiple-Choice options. 
Additionally, UMAE offers flexibility in output and can generate explanations for datasets without explanations for training, e.g. OK-VQA, while also improving answer quality. 
Through case studies and error analysis, we identify potential areas for future improvement, including dataset quality.

\section*{Limitations}
We discuss the limitations of our work in the following two aspects. 
Firstly, the experiments with our proposed framework and finetuning approach are primarily on the OFA model.
We believe our approach applies to any multimodal generative model, however, it would also provide insights to experiment with more models.
Secondly, regarding the evaluation of our proposed joint framework, to better evaluate the generated explanation quality, especially to evaluate the difference between explanations generated jointly with answers and generated conditioned on the answers, human judgement would be an important criterion compared to automatic NLG metrics.

\section*{Acknowledgements}
We acknowledge the support of Apoorv Khandelwal from AI2 for providing us with results for the evaluation of our model predictions over a hidden test set. This was valuable for our earlier draft of the paper. We would like to thank the anonymous reviewers who provided valuable feedback on the previous draft of our paper. 

\bibliography{anthology}
\bibliographystyle{acl_natbib}

\appendix

\section{Datasets} \label{sec:datasets}

The datasets used in the paper are as follows:

\paragraph{OK-VQA} \cite{marino2019ok} is a knowledge-based VQA dataset that requires outside knowledge beyond the images to answer the questions.
It has train and test splits of size 9,009 and 5,046.
Each question is provided answers by five annotators.
To use the VQA \cite{antol2015vqa} metric, each annotated answer is then repeated twice to form a gold answer set with 10 answers.
Since no explanation is provided, we only train \texttt{Q$\rightarrow$A} task on OK-VQA.

\paragraph{A-OKVQA}
\cite{Schwenk2022AOKVQAAB} is currently the largest knowledge-based VQA dataset split into 17.1K, 1.1K, and 6.7K for train, validation, and test, respectively.
The questions cover four knowledge types: visual, commonsense, knowledge bases, and physical.
For each question, it provides both multiple-choice answers and 10 free-form answers (annotated by 10 different people), as well as three explanations. 
Images in both OK-VQA and A-OKVQA are from MSCOCO \cite{lin2014microsoft}, and answers in both datasets are in single words or short phrases.

\paragraph{VCR}
\cite{zellers2019vcr} is a large multiple-choice dataset for Visual Commonsense Reasoning. 
The train, validation, and test splits have 191.6k, 21.3k, and 26.5k instances, respectively.
Each question has four answer options in sentences, and the correct answer is further provided with four explanation options.
Images in VCR are from movie clips \cite{rohrbach2017movie}.
Bounding boxes of entities are provided associated with mentions such as \texttt{Person1} in questions, answers and explanations.
We follow \citet{NEURIPS2021_c6d4eb15} and draw coloured highlights around the referenced entity on the images, where entity names and the coloured highlights are consistent in the entire dataset, expecting the model to learn the association between the coloured bounding box and the entity.

\paragraph{VQA-X}
\cite{park2018multimodal} contains a subset from the VQAv2 \cite{goyal2017making} dataset and further provides three explanations for each question. 
The image-question pairs are split into train, validation, and test with 29.5k, 1.5k, and 2k instances, respectively. 
We only use the original test set to evaluate the zero-shot performance of the trained models.





\section{Hyper-Parameters and Training} \label{sec:parameters}
We begin with the pretrained weights from the original OFA-large\footnote{\url{https://github.com/OFA-Sys/OFA}}, which is trained on vision-only tasks including Image Classification,
language-only tasks including Sentence Classification, Text Summarisation, as well as various vision-language tasks including Image Captioning, Visual Question Answering and Visual Entailment. 
Adam is used as the optimizer and cross-entropy is the loss function.
We set the learning rate to \(10^{-5}\), the warm-up ratio to 0.4, and the patch image size to 480. 
We shuffle all the training examples and use batch size 16.
Due to the large size of VCR, we train for 30 epochs on models involving VCR (\textsc{ofa\textsubscript{q->a}} for VCR, \textsc{umae\textsubscript{vcr}} and \textsc{umae\textsubscript{all}}), and up to 100 epochs for other models.
We report the empirical performance with checkpoints that perform best on the validation set (the 5\% split from the original train set).
For A-OKVQA, we additionally report the answer accuracy on the original test set.


\begin{table}[t!]
\centering
\scalebox{0.8}{
\begin{tabular}{ccc|c}
\toprule
   \sc Question & \sc Objects & \sc Images & \sc Accuracy \\
   \midrule
\sc \cmark     & \cmark   & original    & 50.39        \\
\sc \cmark   & \xmark  & \xmark      & 39.16         \\
\sc \cmark    & \xmark  & random    & 33.48        \\
\sc \cmark    & \cmark  & \xmark    & 33.28        \\
\bottomrule

\end{tabular}
}
\caption{Ablation on the modality dependency for answer accuracy of A-OKVQA.
}
 \label{tab:modality}
\end{table}

\section{Ablations on Modality Dependency}
\label{sec:modality}

 We conduct several ablation studies to investigate the dependency of object features and images on the performance of our model \textsc{UMAE\textsubscript{all}} for answer accuracy of A-OKVQA, where we removed images, replaced them with random images, and removed extracted attributes and features.
Results in \autoref{tab:modality} show that the visual encoder is crucial for performance and that visual objects alone are not sufficient for answer prediction.
Using a random image would introduce noise and therefore performs worse than not including the image at all.
 We did not test removing the question because we believe the model needs the questions to be able to provide answers.

\begin{table*}[t!]
\centering
\scalebox{0.75}{
\addtolength{\tabcolsep}{-1pt}
\begin{tabular}{ll|cccrrccrcc}
\toprule
{\multirow{2}{*}{\sc Dataset}}
 & {\multirow{2}{*}{\sc Decoding}} & {e{\sc-V}i{\sc l }} & \multicolumn{8}{c} {\sc {N-gram} Scores} & \sc Learnt Sc.
\\

&   & \sc \textit{S}\textsubscript{E}     &{\sc bleu\small{1}} & {\sc bleu\small{2}} & \multicolumn{1}{c}{\sc bleu\small{3}} & \multicolumn{1}{c}{\sc bleu\small{4}}  & \sc rouge-l   &\sc meteor  & {\sc cide}r & \sc spice & \sc bertscore    \\\cmidrule(lr){1-2} 
\cmidrule(lr){3-3} \cmidrule(lr){4-11} \cmidrule(lr){12-12}

{\multirow{4}{*}{\sc {A-okvqa}}}  & \sc beamsearch &  44.71 & 52.01 & 36.69 & 26.72 & 19.88 & 40.39 & 22.06 & 68.48 & 20.94   & 86.05  \\
&  {\sc  top-k} ($k=100$) & 44.34 & 52.56 & 37.06 & 27.06 & 19.72 & 44.45 & 21.58 & 73.44 & 19.38  & 86.27\\
& {\sc  nucleus} ($p=0.4$) &\textbf{50.82} & 58.92 & 44.66 & 35.06 & 27.35 & 52.56 & 24.83 & 101.09 & 23.33 & 88.21 \\
&{\sc typical} ($p=0.6$) & 47.27 & 54.18 & 39.39 & 29.82 & 22.18 & 47.78 & 22.79 & 84.43 & 21.47  & 86.95\\
\midrule
{\multirow{4}{*}{\sc {Vcr}}} & \sc beamsearch  & 40.23 & 26.41 & 20.15 & 15.95& 12.47 & 29.13 & 16.82 & 49.72  & 27.70 & 81.84\\
&  {\sc top-k} ($k=50$) & 33.19 & 20.98 & 14.89 & 11.18 & 8.33 & 23.65 & 13.72 & 32.73 & 21.99 & 80.31 \\
& {\sc nucleus} ($p=0.1$) & \textbf{40.27} & 31.42 & 22.95 & 17.62 & 13.44 & 29.53 & 17.54 & 47.33 & 26.45   & 81.91 \\
&{\sc typical} ($p=0.4$) &35.12 & 23.42 & 16.88 & 12.83 & 9.64  & 25.36 & 14.70 & 35.85 & 23.32  & 80.70 \\
\midrule
{\multirow{4}{*}{\sc {Vqa-x}}} & \sc beamsearch & \cellcolor[HTML]{DDEBF7}35.88     & \cellcolor[HTML]{DDEBF7}37.84& \cellcolor[HTML]{DDEBF7}24.91 & \cellcolor[HTML]{DDEBF7} 16.67& \cellcolor[HTML]{DDEBF7}10.97       & \cellcolor[HTML]{DDEBF7}31.32       &\cellcolor[HTML]{DDEBF7}17.90     &\cellcolor[HTML]{DDEBF7}38.23   & \cellcolor[HTML]{DDEBF7}16.23      &\cellcolor[HTML]{DDEBF7}84.39\\
&  {\sc top-k} ($k=50$) & \cellcolor[HTML]{DDEBF7}33.28 &\cellcolor[HTML]{DDEBF7}38.35 & \cellcolor[HTML]{DDEBF7}23.11 &\cellcolor[HTML]{DDEBF7}14.21 &\cellcolor[HTML]{DDEBF7}8.45  &\cellcolor[HTML]{DDEBF7}29.15 &\cellcolor[HTML]{DDEBF7}17.05 & \cellcolor[HTML]{DDEBF7}32.89 & \cellcolor[HTML]{DDEBF7}15.26 &\cellcolor[HTML]{DDEBF7}83.41\\
&{\sc nucleus} ($p=0.1$) & \cellcolor[HTML]{DDEBF7}\textbf{40.67} &\cellcolor[HTML]{DDEBF7}47.56 & \cellcolor[HTML]{DDEBF7}31.44 & \cellcolor[HTML]{DDEBF7}21.47 & \cellcolor[HTML]{DDEBF7}14.63 &\cellcolor[HTML]{DDEBF7}35.12 & \cellcolor[HTML]{DDEBF7}20.29 &\cellcolor[HTML]{DDEBF7}50.35 & \cellcolor[HTML]{DDEBF7}19.13  &\cellcolor[HTML]{DDEBF7}85.40\\
&{\sc typical} ($p=0.5$) & \cellcolor[HTML]{DDEBF7}36.31 &\cellcolor[HTML]{DDEBF7}40.85 &\cellcolor[HTML]{DDEBF7}25.57 &\cellcolor[HTML]{DDEBF7}16.82 &\cellcolor[HTML]{DDEBF7}11.14 &\cellcolor[HTML]{DDEBF7}31.08 &\cellcolor[HTML]{DDEBF7}18.15 &\cellcolor[HTML]{DDEBF7}39.71 &\cellcolor[HTML]{DDEBF7}16.62  &\cellcolor[HTML]{DDEBF7}83.93 \\
\bottomrule
\end{tabular}

}
\caption{Explanation scores with automatic NLG for generated explanations (\texttt{QA$\rightarrow$E}) from \textsc{umae\textsubscript{all}} model with different decoding strategies.
The last two rows (with blue shadow) indicate out-of-domain performance.
}
\label{tab:sampling}
\end{table*}

\begin{table*}[ht]
\centering
\scalebox{0.75}{
\addtolength{\tabcolsep}{-1pt}
\begin{tabular}{ll|cccrrccrcc}
\toprule
{\multirow{2}{*}{\sc Dataset}}
 & {\multirow{2}{*}{\sc Decoding}} & {e{\sc-V}i{\sc l }} & \multicolumn{8}{c} {\sc {N-gram} Scores} & \sc Learnt Sc.
\\

&   & \sc \textit{S}\textsubscript{E}      &{\sc bleu\small{1}} & {\sc bleu\small{2}} & \multicolumn{1}{c}{\sc bleu\small{3}} & \multicolumn{1}{c}{\sc bleu\small{4}}  & \sc rouge-l   &\sc meteor & {\sc cide}r & \sc spice & \sc bertscore    \\\cmidrule(lr){1-2} 
\cmidrule(lr){3-3} \cmidrule(lr){4-11} \cmidrule(lr){12-12}

{\multirow{2}{*}{\sc {A-okvqa}}}  & \sc beamsearch &  \textbf{47.01} & 54.75 & 41.39 & \textbf{32.08} & \textbf{24.25} & \textbf{49.75} & \textbf{22.54} & \textbf{86.28} & \textbf{20.68} & \textbf{87.39}  \\
& {\sc  nucleus} ($p=0.5$) &46.72   & \textbf{55.53} & \textbf{41.63} & 31.91   & 23.67  & 49.16   & 22.48  & 82.37  & 20.67   & 87.18    \\
\midrule
{\multirow{2}{*}{\sc {Vcr}}} & \sc beamsearch  &\textbf{37.02} & 25.00          & 18.90          & \textbf{14.87} & \textbf{11.54} & \textbf{27.07} & \textbf{15.66} & \textbf{38.77} & \textbf{25.03} & \textbf{80.68} \\
& {\sc nucleus} ($p=0.1$) &  35.10     & \textbf{27.41} & \textbf{19.36} & 14.50   & 10.73  & 26.18    & 15.21 & 34.99    & 21.88  & 80.52  \\

\midrule
{\multirow{2}{*}{\sc {Vqa-x}}} & \sc beamsearch & \cellcolor[HTML]{DDEBF7}38.13 & \cellcolor[HTML]{DDEBF7}39.91 & \cellcolor[HTML]{DDEBF7}26.30 & \cellcolor[HTML]{DDEBF7}17.99 & \cellcolor[HTML]{DDEBF7}12.46 &\cellcolor[HTML]{DDEBF7}31.69 & \cellcolor[HTML]{DDEBF7}19.11 & \cellcolor[HTML]{DDEBF7}42.10 & \cellcolor[HTML]{DDEBF7}18.15 & \cellcolor[HTML]{DDEBF7}84.95\\
&{\sc nucleus} ($p=0.1$) & \cellcolor[HTML]{DDEBF7}\textbf{39.67} & \cellcolor[HTML]{DDEBF7}\textbf{44.92} & \cellcolor[HTML]{DDEBF7}\textbf{28.88} & \cellcolor[HTML]{DDEBF7}\textbf{19.04} & \cellcolor[HTML]{DDEBF7}\textbf{12.55} & \cellcolor[HTML]{DDEBF7}\textbf{33.08} &\cellcolor[HTML]{DDEBF7}\textbf{20.07} &\cellcolor[HTML]{DDEBF7}\textbf{44.28} & \cellcolor[HTML]{DDEBF7}\textbf{19.19} & \cellcolor[HTML]{DDEBF7}\textbf{85.21}\\

\bottomrule
\end{tabular}
}
\caption{Explanation scores with automatic NLG for generated explanations from \texttt{Q$\rightarrow$AE} with \textsc{umae\textsubscript{all}} model.
The last two rows (with blue shadow) indicate out-of-domain performance.
}
\label{tab:ar_detail}
\end{table*}

\begin{figure*}[ht]
\centering
    \includegraphics[width=\linewidth]{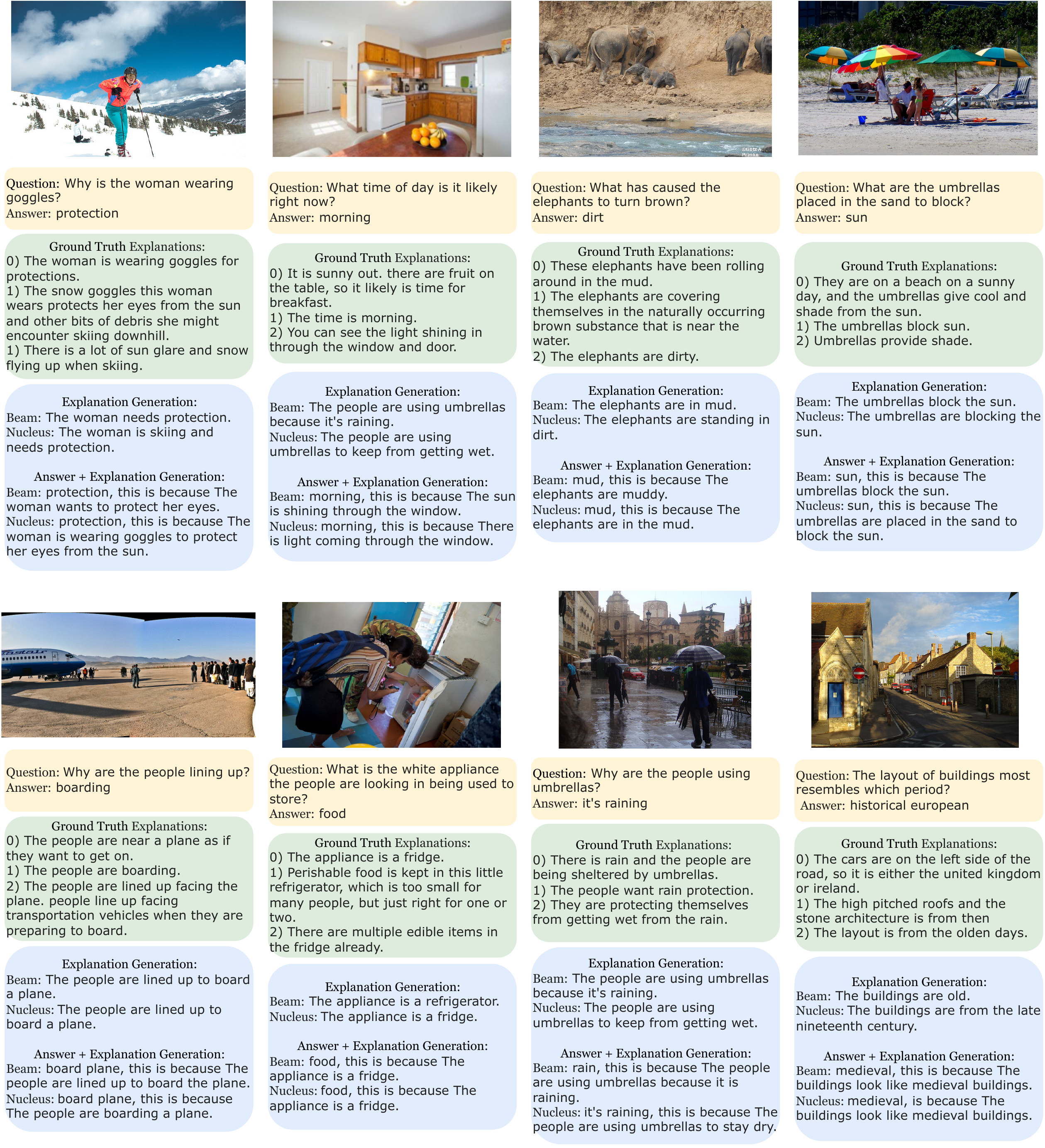}
    \caption{Examples of generated answers and explanations for A-OKVQA.}
\label{fig:aok}
\end{figure*} 

\begin{figure*}[ht]
\centering
    \includegraphics[width=\linewidth]{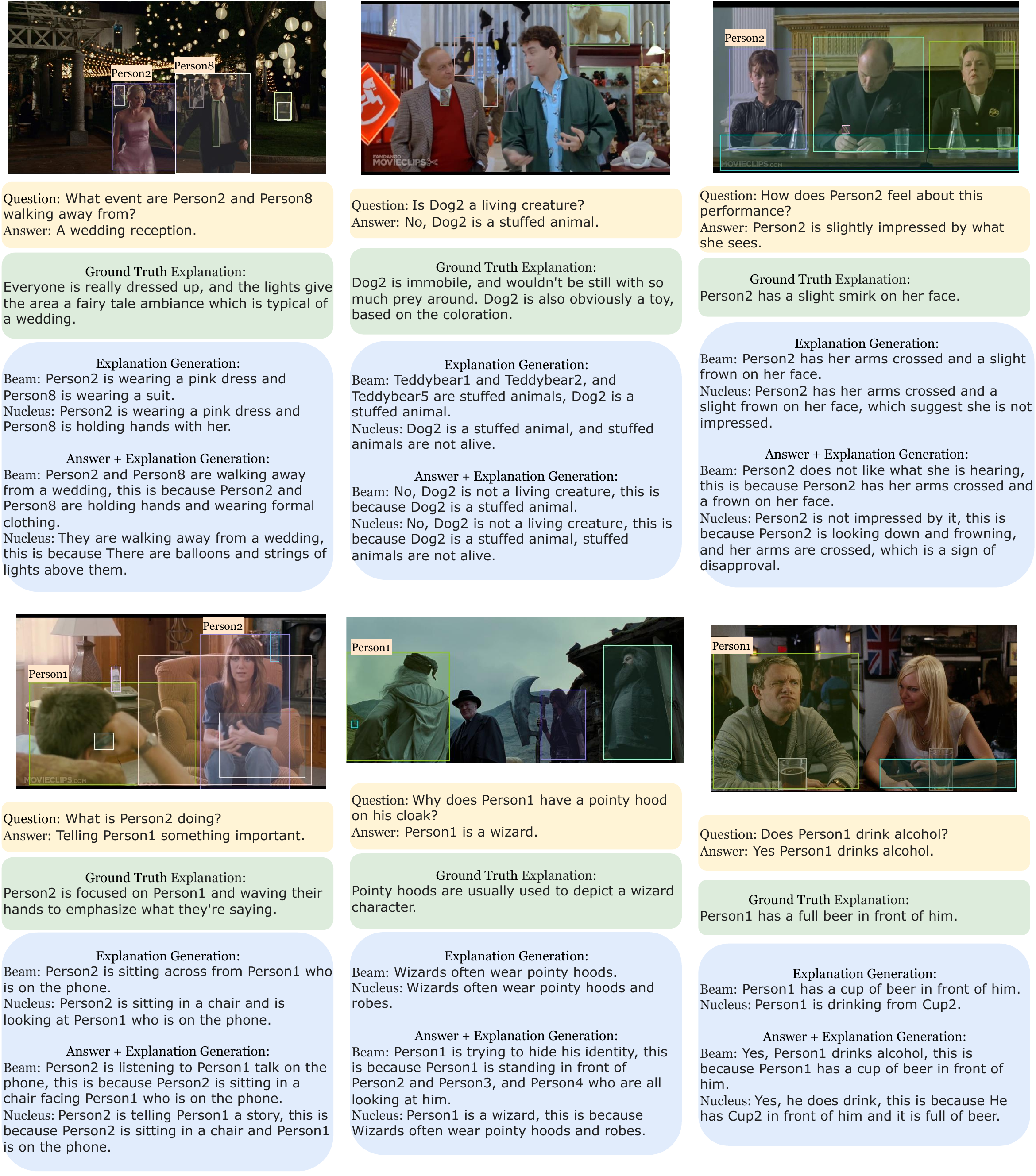}
    \caption{Examples of generated answers and explanations generation for VCR.}
\label{fig:vcr}
\end{figure*}

\section{More Explanation Scores} \label{sec:explanation_scores}
For decoding, we evaluate the performance of beam search with the size of 5, top-k sampling with $k$ from $\{50, 100, 200, ..., 1000\}$, and Nucleus and Typical \cite{meister2022typical} sampling, both with $p$ from $\{0.1, 0.2, ..., 0.9\}$.
We show the details of the NLG scores using different decoding strategies for explanations generated from \texttt{QA$\rightarrow$E} in \autoref{tab:sampling}, and \texttt{Q$\rightarrow$AE} in \autoref{tab:ar_detail}.

\section{Examples of Generated Explanations} \label{sec:explanations}
Examples of the explanations generated with beam search and Nucleus sampling for A-OKVQA are shown in \autoref{fig:aok}, and VCR in \autoref{fig:vcr}.

\section{Joint Generation Performance} \label{sec:joint}

We present the results of the proposed \texttt{Q$\rightarrow$AE} task where answers and explanations are jointly generated. 
We parse the generated sequence to the answer and the explanation and use the same sets of metrics as the separate generation for evaluation. 
Results for answers in \autoref{tab:AR_ans}  and explanations in \autoref{tab:AR}.
For answers, since the perplexity metric does not directly compare the generation, we show the Multiple-Choice accuracy using the Glove metric for A-OKVQA and \texttt{BERTScore} for VCR answer sentences.

\begin{table}[!t]
\centering
\scalebox{0.8}{
\addtolength{\tabcolsep}{-1pt}
\begin{tabular}{l|cccccc}
\toprule
\multirow{2}{*}{\sc Task} & {\sc A-okvqa} & {\sc Vcr} &  {\sc Vqa-x}    \\
\cmidrule(lr){2-2} \cmidrule(lr){3-3} \cmidrule(lr){4-4} 
& \sc mc (golve) & \sc bertscore & \sc da\\

\midrule
\sc{q}{\small{->}}{\sc a} & 65.67 & 81.91  & \cellcolor[HTML]{DDEBF7}77.65\\
\sc{q}{\small{->}}{\sc ae} & 65.67 & 82.30 & \cellcolor[HTML]{DDEBF7}69.60\\

\bottomrule
\end{tabular}
}
\caption{Evaluation of answers generated given questions (\texttt{Q->A}) and jointly generated with explanations (\texttt{Q->AE}).
MC stands for Multiple Choice, DA for Direct Answer.
The last column with a blue shadow indicates out-of-domain performance.}
\label{tab:AR_ans}
\end{table}

\begin{table}[!t]
\centering
\scalebox{0.75}{
\addtolength{\tabcolsep}{-1pt}
\begin{tabular}{l|cccccc}
\toprule
\multirow{2}{*}{\sc Dataset} & \multicolumn{2}{c}{\sc \textit{S}\textsubscript{E}}   & \multicolumn{2}{c}{\sc Ngramscore}   &  \multicolumn{2}{c}{\sc Bertscore}    \\
\cmidrule(lr){2-3} \cmidrule(lr){4-5} \cmidrule(lr){6-7}
& \sc{qa}{\small{->}}{\sc e} &  \sc{q}{\small{->}}{\sc ae} & \sc{qa}{\small{->}}{\sc e} &  \sc{q}{\small{->}}{\sc ae} & \sc{qa}{\small{->}}{\sc e} &  \sc{q}{\small{->}}{\sc ae} \\
\midrule
\sc {A-okvqa}& 50.82 & 47.01 & 35.69 & 32.15 & 88.21 & 87.39  \\

\sc {Vcr}& 40.27 & 37.02 &  26.70 &24.02 & 81.91 & 80.68\\
\sc {Vqa-x}  & \cellcolor[HTML]{DDEBF7}40.67 &\cellcolor[HTML]{DDEBF7}39.67 & \cellcolor[HTML]{DDEBF7}26.69 & \cellcolor[HTML]{DDEBF7}25.85 & \cellcolor[HTML]{DDEBF7}85.40 &\cellcolor[HTML]{DDEBF7}85.21 \\

\bottomrule
\end{tabular}
}
\caption{Scores of explanations generated given answers (\texttt{QA->E}) and jointly generated with answers (\texttt{Q->AE}).
The last row with a blue shadow indicates out-of-domain performance.}
\label{tab:AR}
\end{table}

\begin{table}[t!]
\centering
\scalebox{0.8}{
\begin{tabular}{l|ccc}
\toprule
     & \sc Ok-vqa     & \multicolumn{2}{c}{\sc A-okvqa}           \\
      & \sc da  & \sc mc (glove) & \sc da \\ 
       \cmidrule(lr){1-1} \cmidrule(lr){2-2} \cmidrule(lr){3-4}
\sc Best     & 80.94  & 80.74    & 66.20        \\
\sc Average  & 54.98  & 71.53      & 57.29         \\
\sc Worst   & 16.37 & 59.35    & 41.46        \\
\bottomrule

\end{tabular}
}
\caption{Human performance on OK-VQA and A-OKVQA measured from the ground truth answers.
}
 \label{tab:human}
\end{table}

\section{Datasets Quality and Issues} \label{sec:datasets_issues}
As mentioned in \autoref{sec:analysis}, during error analysis we found that many errors are due to the issue in the dataset itself. 
Concretely, we observe the following issues in the existing datasets:
(1) wrong answers
(2) subjective or unanswerable questions
(3) typos or unclear expressions
(4) not requiring images or knowledge to answer the question as designed.

Furthermore, since the answer and explanation for a question in VCR are obtained from the same person who authored the question, this may result in severe subjectivity in the answers or explanations.
For example, we find that many questions in VCR require knowledge of the \textit{movie plot} from which the image is extracted, rather than \textit{commonsense reasoning} to answer the questions. 
While human annotators have an implicit understanding of the movies, the dataset itself does not contain relevant contextual information. 

\begin{figure*}[ht]
\centering
    \includegraphics[width=0.85\linewidth]{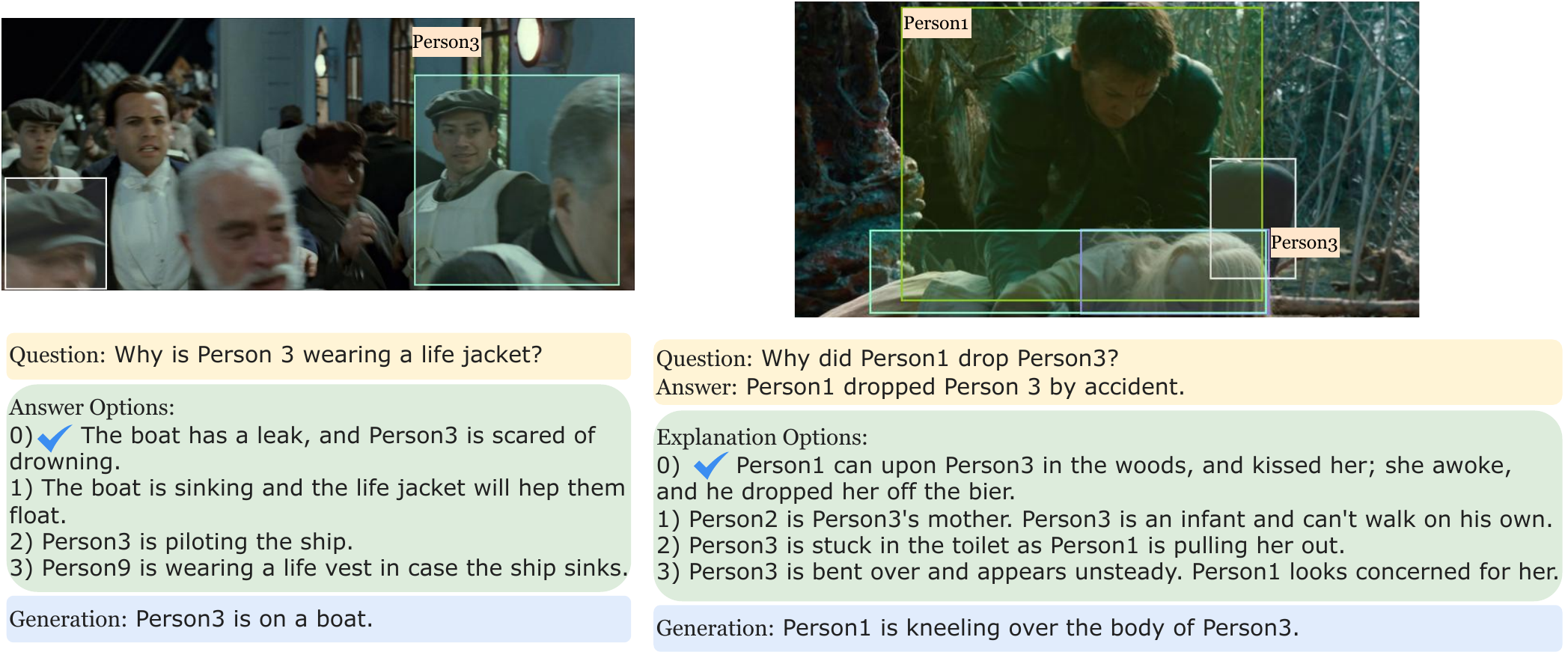}
    \caption{Questions that require knowledge of the movie plots to generate the answers from VCR.}
\label{fig:moive_know}
\end{figure*}

\begin{figure*}[ht]
\centering
    \includegraphics[width=0.85\linewidth]{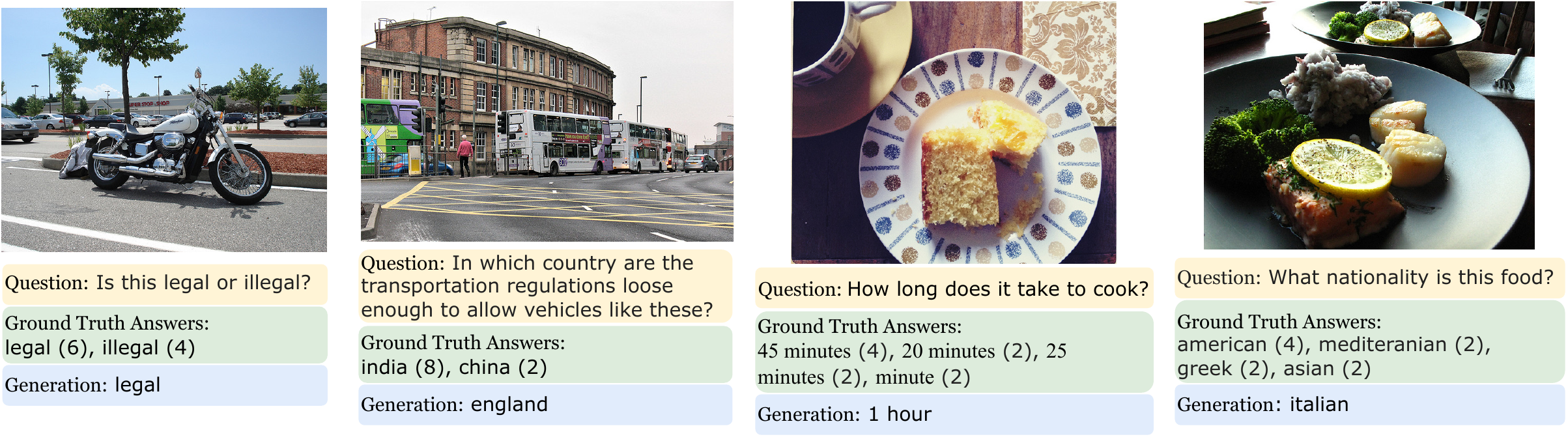}
    \caption{Examples of subjective questions from OK-VQA.}
\label{fig:dataset}
\end{figure*}

\begin{figure*}[ht]
\centering
    \includegraphics[width=0.85\linewidth]{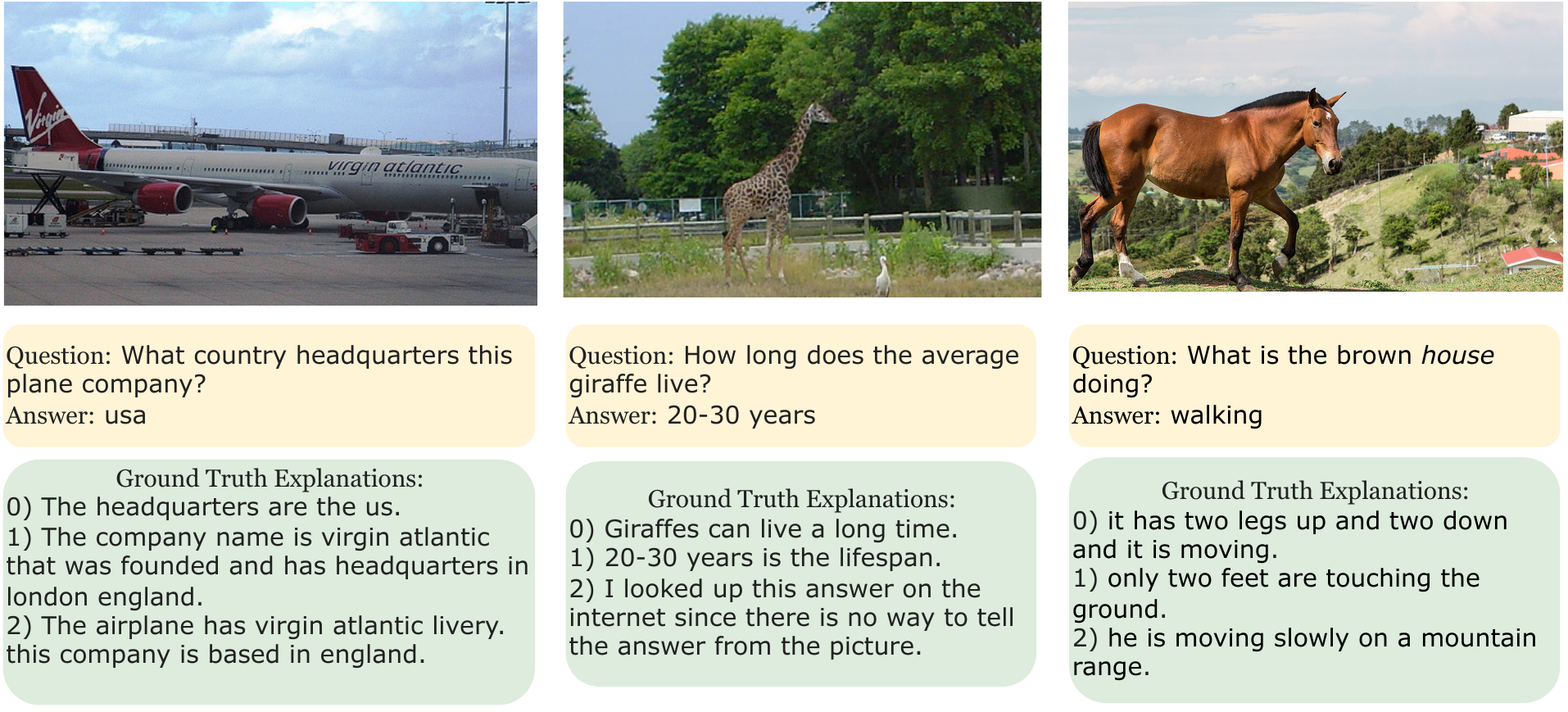}
    \caption{Issues in the datasets that severely impact the model generation: wrong answers (left, from A-OKVQA), questions do not need visual input to answer (middle, from A-OKVQA), and typo (right, from VQA-X).}
\label{fig:dataset_issues}
\end{figure*}

We show some of the issues in the datasets below.
\autoref{fig:moive_know} shows examples from VCR that require an understanding of the movie plot to generate answers.
\autoref{fig:dataset} shows examples from OK-VQA where questions and answers are subjective or ambiguous.
\autoref{fig:dataset_issues} shows examples from A-OKVQA and VQA-X that either contain wrong answers, questions that do not need visual input or typos which severely impact the model generation (``house'' should be ``horse''). 

To understand the inter-annotator agreement for the datasets, we further measure the best, average and worst human performance on OK-VQA and A-OKVQA by selecting the most common answer, a random answer, and the least common answer, respectively, from the 10 ground truth answers for each question.
We calculate the performance using the VQA metric for direct answers, and the GloVe metric for Multiple Choice for simplicity.
Note that we also remove the answer selected from the ground truth answers when measuring human performance.
From the results in \autoref{tab:human} we can see that the average performance on both datasets is relatively poor, which indicates the noise in the datasets.
The quality of the datasets needs to be more carefully inspected so that the model performance evaluated on these datasets can be more meaningful.

\end{document}